\pdfoutput=1

\documentclass[10pt,twocolumn,letterpaper]{article}
\usepackage[accsupp]{axessibility} 

\usepackage{wacv}
\usepackage{times}
\usepackage{epsfig}
\usepackage{graphicx}
\usepackage{amsmath}
\usepackage{amssymb}
\usepackage{booktabs}

\usepackage{multirow}
\usepackage{caption}
\usepackage{subcaption}
\usepackage{tikzscale}
\usepackage{pgfplots,pgfplotstable}

\def\wacvPaperID{0749} 

\wacvapplicationstrack 

\wacvfinalcopy 

\ifwacvfinal
\usepackage[breaklinks=true,bookmarks=false]{hyperref}
\else
\usepackage[pagebackref=true,breaklinks=true,colorlinks,bookmarks=false]{hyperref}
\fi

\begin{document}

\title{Adaptive Feature Fusion for Cooperative Perception using LiDAR Point Clouds}

\author{Donghao Qiao, Farhana Zulkernine\\
Queen’s University, Canada\\
{\tt\small \{d.qiao, farhana.zulkernine\}@queensu.ca}
}

\maketitle
\thispagestyle{empty}

\begin{abstract}
Cooperative perception allows a Connected Autonomous Vehicle (CAV) to interact with the other CAVs in the vicinity to enhance perception of surrounding objects to increase safety and reliability. It can compensate for the limitations of the conventional vehicular perception such as blind spots, low resolution, and weather effects. An effective feature fusion model for the intermediate fusion methods of cooperative perception can improve feature selection and information aggregation to further enhance the perception accuracy. We propose adaptive feature fusion models with trainable feature selection modules. One of our proposed models Spatial-wise Adaptive feature Fusion (S-AdaFusion) outperforms all other State-of-the-Arts (SOTAs) on two subsets of the OPV2V dataset: Default CARLA Towns for vehicle detection and the Culver City for domain adaptation. In addition, previous studies have only tested cooperative perception for vehicle detection. A pedestrian, however, is much more likely to be seriously injured in a traffic accident. We evaluate the performance of cooperative perception for both vehicle and pedestrian detection using the CODD dataset. Our architecture achieves higher Average Precision (AP) than other existing models for both vehicle and pedestrian detection on the CODD dataset. The experiments demonstrate that cooperative perception also improves the pedestrian detection accuracy compared to the conventional single vehicle perception process.

\end{abstract}

\section{Introduction}


3D object detection with LiDAR sensor has become more significant in Autonomous Vehicles (AVs) in the recent years \cite{voxelrcnn,pointpillars,pvrcnn++,second,voxelnet}, because it can provide more spatial information about the objects of interest including their locations, size and orientation. LiDAR can generate point cloud data which contains accurate depth information and is less affected by external illumination conditions. However, point clouds far away from the LiDAR are extremely sparse which make the detection of further objects more difficult. Objects that are occluded will generate fewer points, which makes inference harder especially for the small objects such as pedestrians.

Cooperative perception enables Connected Autonomous Vehicles (CAVs) and/or Road Side Units (RSUs) to share perceived information using the Vehicular Communication (VC) systems within the communication range \cite{infrastructurecooper,fcooper,sumcooper,v2vnet,cobevt,v2xvit,opv2v}. The perceived information can consist of GPS and a variety of sensor data including RADAR, camera, and LiDAR data. Cooperative perception helps to compensate for the limitations of the current visual perception techniques such as limited resolution, weather effects, and blind spots. CAVs are equipped with VC systems to enable exchange of traffic information with the surrounding CAVs. One vehicle can receive and aggregate information from other CAVs with its own locally perceived data to implement cooperative perception and thereby, improve the safety and robustness of the automated driving system. 

\begin{figure}[t]
\centering
\includegraphics[width=0.99\columnwidth]{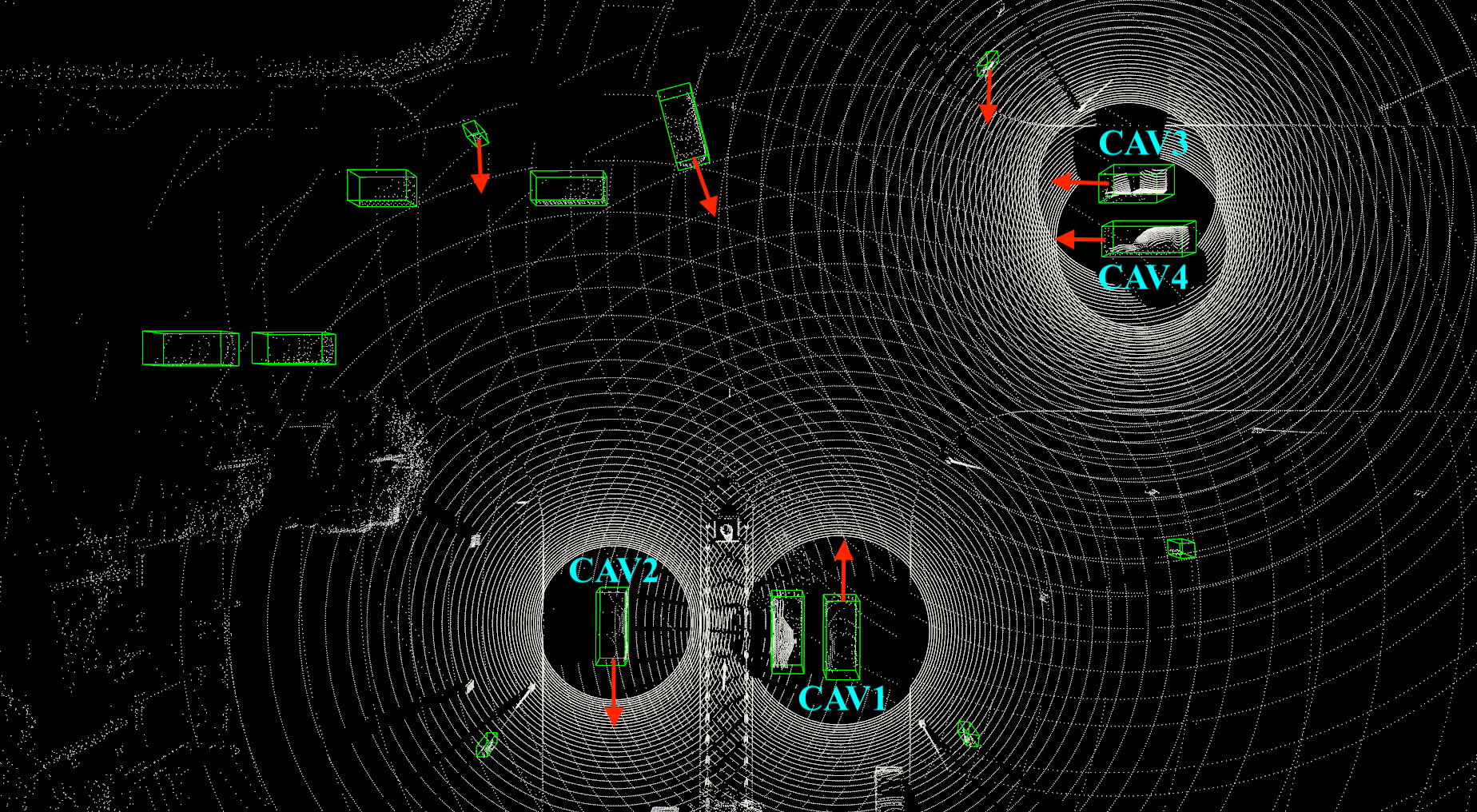}
\caption{One example frame from CODD dataset \cite{codd} with four CAVs. The red arrows indicate the moving directions of four CAVs, one left turn vehicle and two pedestrians.}
\label{fig:case1}
\end{figure}

An example from the CODD dataset \cite{codd} is shown in Fig. \ref{fig:case1}. The figure depicts a scene at an intersection with four CAVs. In this scenario, CAV1 is driving towards the horizontal main street and will interact with other vehicles and pedestrians. Even if CAV2 is leaving the main street, it can still provide other CAVs with the view of the intersection. At the same time, CAV3 and CAV4 will supply CAV1 with the main street view, allowing it to pass the intersection safely and efficiently. Furthermore, when CAV4's vision is obstructed by CAV3 while the former is detecting pedestrian crossing, CAV1 and CAV3 will still be able to share pedestrian crossing information with CAV4. Suppose CAV1 is selected as the ego vehicle and all predictions are driven by the ego vehicle. First, the four CAVs process their LiDAR point clouds and extract the intermediate feature maps in parallel in their local system. Next, the other three CAVs broadcast their extracted feature maps to CAV1 along with the LiDAR pose information. Then, CAV1 projects the three feature maps to its own coordinate system and aggregates the information with its own perceived information for 3D object detection. 

According to the data type that is shared among the CAVs, three kinds of data fusion methods are found in the existing literature: 1) early fusion \cite{cooper} aggregates the raw input sensor data from other CAVs; 2) intermediate fusion \cite{fcooper,sumcooper,v2vnet,opv2v} aggregates the processed feature maps from other CAVs; and 3) late fusion \cite{trupercept,zhang2021distributed} aggregates the predicted outputs of object detection from other CAVs. In recent studies, the intermediate feature fusion has repeatedly proved to be the most efficient fusion method compared to the early and late fusion methods. We hypothesize that the intermediate fusion method can be improved further for real time perception and greater accuracy by implementing an effective feature selection and fusion module having reduced computational cost.

The Vehicle-to-Vehicle (V2V) communication allows vehicles to communicate over a wireless network having a low bandwidth and limited communication range. Because of this, efficient feature extraction, data dissemination, and data fusion are essential for cooperative perception. The PointPillars \cite{pointpillars} backbone is applied for encoding point cloud owing to its short inference time and comparable accuracy. It runs at 62 Hz with 59.20\% mAP on the KITTI \cite{kitti} dataset. A feature extraction module with 2D Convolutional Neural Networks (CNNs) is utilized to further extract features from the encoded point cloud. AVs must demonstrate extremely high accuracy and efficiency. With these prerequisites, we propose several feature fusion models with trainable neural network that can adaptively select features from multiple CAVs.

\paragraph{Contribution.} The contribution of this work are as follows: 1) we create a lightweight cooperative perception architecture with intermediate fusion; 2) we involve 3D CNN and adaptive feature fusion for cooperative perception and propose three trainable feature fusion models for cooperative perception; 3) we validate the proposed models using two public cooperative perception benchmark datasets (OPV2V dataset \cite{opv2v} and CODD dataset \cite{codd}) for multiple tasks including a) vehicle detection, b) pedestrian detection and c) domain adaptation; 4) we experiment with different number of CAVs to observe its influence on cooperative perception.

Our S-AdaFusion model outperforms all the existing models on the OPV2V Default CARLA Towns for the vehicle detection task, and on the OPV2V Culver City for the domain adaptation task. We also validate our model on the CODD dataset for vehicle and pedestrian detection tasks. Our model achieves higher Average Precision (AP) than all other state-of-the-art (SOTA) models. Our experiments also demonstrate that cooperative perception with LiDAR point clouds can improve the accuracy of pedestrian detection compared to the conventional perception process.

The paper is organized as follows. Section \ref{sec:related_work} describes the related work on cooperative perception and feature fusion models. Our proposed cooperative perception framework and feature fusion models are illustrated in Section \ref{sec:framework} and Section \ref{sec:methodology} respectively. The experimental results are presented in Section \ref{sec:experiments}. Section \ref{sec:conclusion} concludes the paper with a list of future works.

\section{Background and Related Work} \label{sec:related_work}
We first present the formal description of cooperative perception and the relevant feature integration methods as background concepts. Then the related works are discussed to highlight the research gap with regard to our research contribution. 

\subsection{Cooperative Perception} \label{sec:cooperative perception}

Formally, the problem of cooperative perception can be described as follows. We denote the raw input data (camera data and LiDAR data) as $I=\{I_1, I_2, \ldots, I_n\}$ from a set of surrounding CAVs as $V=\{v_1, v_2, \ldots, v_n\}$. The corresponding set of extracted intermediate features for the input data is represented as $F=\{F_1, F_2, \ldots, F_n\}$, where the predicted set of outputs from object detectors is denoted as $O=\{O_1, O_2, \ldots, O_n\}$. In the traditional visual perception process, an AV $v_i$ receives raw data $I_i$ from sensors such as camera and LiDAR. This data is then processed to extract a feature map $F_i$ to be used in computational models for predicting objects as output $O_i$. In cooperative perception, an additional data fusion step is applied to aggregate the data from other vehicles for improved perception.

Researchers have demonstrated three kinds of data fusion methods (early fusion, intermediate fusion and late fusion) based on the types of information in the data dissemination stage. CAVs broadcast raw sensor data $I$ in early fusion \cite{cooper} which incurs highest data transfer cost. Late fusion \cite{trupercept, zhang2021distributed} shares and aggregates the predictions $O$ from the CAVs and causes less burden on data transfer, but the object detection performance highly relies on the other CAVs' prediction accuracy and the post-processing of the predictions. The performance of cooperative perception with early and late fusion can be improved by optimizing the 3D object detectors and post-processing method \cite{zhang2021distributed}. Intermediate fusion is a compromise that leverages the processed intermediate feature representations $F$. Therefore, accurate and optimized integration and processing of the information obtained from different locations becomes critical for effective feature fusion to enable accurate object detection. Marvasti et al. \cite{sumcooper} warp the 3D LiDAR point clouds into Bird's-Eye View (BEV) and apply 2D CNN to extract intermediate features in each connected AV. The feature maps from the CAVs are projected on to the ego vehicle's coordinate system. Then these are aggregated with the ego vehicle's feature map. In \cite{sumcooper}, only two CAVs are utilized and the summation makes the overlaps have larger weights, whereas in real-life scenarios, the numbers of CAVs vary. We compute the mean at the overlaps instead. Chen et al. \cite{fcooper} propose feature-level fusion schemes, and the maximum at the overlaps is selected to represent the intermediate features.

The models mentioned above utilize simple reduction operators such as summation, max pooling or average pooling. These operators are able to process the information at the overlaps and fuse feature maps with negligible computational cost. However, the selected features are not necessarily the best due to lack of information selection and identification of data correlation. In V2VNet \cite{v2vnet}, a Graph Neural Network (GNN) is applied to represent a map of CAVs based on the geological coordinates to facilitate data fusion. The GNN aggregates information received from multiple vehicles with the vehicle’s internal state (computed from its own sensor data) to compute an updated intermediate representation. Xu et al. \cite{opv2v} propose AttFuse and leverage self-attention model \cite{attention} to fuse the intermediate feature maps. Transformers are utilized in V2X-Vit \cite{v2xvit} and CoBEVT \cite{cobevt} for cooperative perception with intermediate feature fusion. We explore the feature fusion models that can utilize multiple feature maps from the CAVs effectively and efficiently.

\subsection{Feature Learning and Feature Fusion} \label{subsec:feature_fusion}

The attention mechanism has demonstrated its utility in solving computer vision tasks \cite{senet,qiao2020dilated,attention,cbam}. By incorporating a small module in the neural network, the model can leverage the channel and/or spatial information, and enhance the extracted representation.

Feature learning and feature fusion have also been explored in the past for camera and LiDAR in 3D object detection. Yoo et al. \cite{3dcvf} proposed an adaptive gated fusion network to combine both LiDAR and image features. Liu et al. \cite{bevfusion1} and Liang et al. \cite{bevfusion2} propose a BEV fusion method that project intermediate features from the images captured using a camera and point clouds generated by the LiDAR into BEV, and concatenate the two feature maps along the channel axes. Liu et al. \cite{bevfusion1} utilize a convolutional encoder to generate a refined fusion based on the BEV feature map. Liang et al. \cite{bevfusion2} propose a dynamic fusion module and apply channel-wise attention to adaptively select features.

Concatenating the intermediate feature maps by the feature channel in cooperative perception can drastically increase the computational cost with an increasing number of CAVs. Therefore, instead of concatenating feature maps and creating an extra large feature extraction network, it is more efficient to aggregate the feature maps with geometric and geological information.

\section{Overview of the Proposed Framework} \label{sec:framework}

\begin{figure*}[htbp]
\centering
\includegraphics[width=0.9\textwidth]{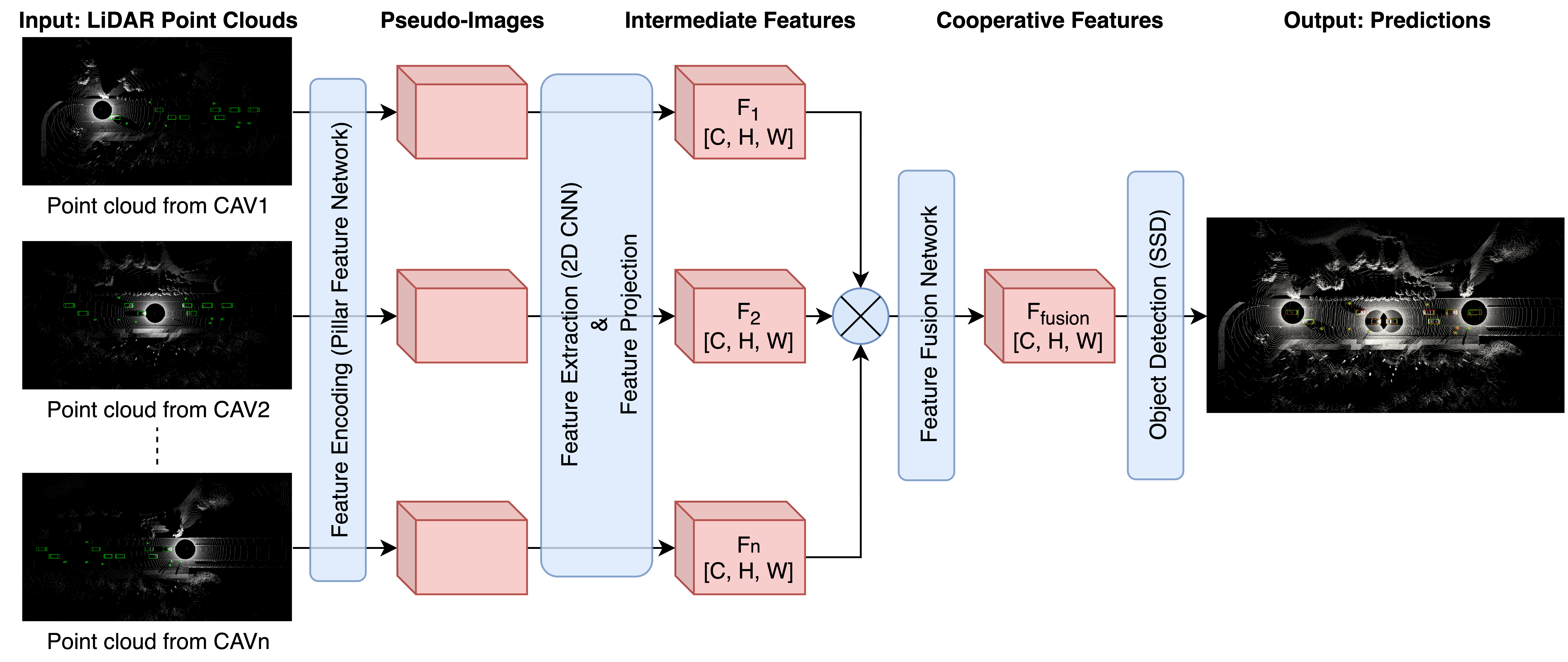}
\caption{Architecture of the cooperative perception model.}
\label{fig:model}
\end{figure*}

The proposed feature fusion based cooperative perception architecture is extended from the PointPillars \cite{pointpillars} as depicted in Fig. \ref{fig:model}. The overall network takes point cloud as input and processes the data in 5 steps: 1) feature encoding converts the point cloud into a pseudo-image with Pillar Feature Network (PFN); 2) intermediate feature extraction extracts multi-scale features from the pseudo-image with a 2D pyramid network; 3) feature projection projects the intermediate feature maps from CAVs on to the ego vehicle coordinate with the LiDAR pose information; 4) intermediate feature fusion generates the combined feature map with a feature fusion network; and 5) 3D object detection regresses the bounding boxes with the Single Shot Detector (SSD) \cite{ssd} and predicts detected objects' classes.

\subsection{Feature Encoding}

The input point cloud with dimension $(n\times4)$ consists of $n$ points, and each point has attributes $(x, y, z)$ coordinates and intensity. The point cloud is encoded into pillars with the height equal to the point cloud height in $z$ axis. The points in each pillar are augmented with 5 extra attributes including the offsets from the arithmetic mean of all points in the pillar and the offsets from the pillar center. The point cloud data is transformed into $P$ pillars with each pillar having $N$ points and $D$ features. Then the PointNet \cite{pointnet} is applied on the pillars to extract features and generate a tensor of size $(C_{in} \times P)$. The pillars with $C_{in}$ features are projected back on to the original location to generate a pseudo-image of size $(C_{in} \times 2H \times 2W)$. We use $2H$ and $2W$ here, since we downsample the feature maps to $(C \times H \times W)$ in the next feature extraction step.

\subsection{Feature Extraction}

Similar to PointPillars \cite{pointpillars}, a Feature Pyramid Network (FPN) \cite{fpn,qiao2021drivable} is utilized to extract intermediate features from the pseudo-images. The FPN contains three downsample blocks for multi-resolution feature extraction. Each block contains a 2D convolution layer, a batch normalization layer and a ReLU activation function. The three feature maps obtained from the three downsample blocks are then upsampled and concatenated. The resulting multiscaled feature map is refined by a CNN to generate a feature map of size $F^{'} \in R^{C \times H \times W}$.

\subsection{Feature Projection}

Feature maps extracted from the different CAVs have different geological locations and orientations. Therefore, they need to be transformed into the receiver's coordinate system for feature fusion and object detection. The CAVs disseminate the feature maps along with their LiDAR pose information containing $(X, Y, Z, roll, yaw, pitch)$. Once the ego vehicle receives data from the neighboring CAVs, it is projected into the ego vehicle’s coordinate system and timestamp. In this work, we use the intermediate feature projection and integration.

\subsection{Feature Fusion}

The projected intermediate feature maps from different CAVs are expanded to 4D tensors and concatenated for further processing. As explained in the related work Section \ref{subsec:feature_fusion}, concatenating the feature maps over the feature channel generates a 3D tensor with $nC$ channels, which will increase computational complexity and cost for feature fusion and refinement. Therefore, we aggregate the feature maps into a 4D tensor $F \in R^{n \times C \times H \times W}$ where $n$ is the maximum number of CAVs. To fuse the overlapping feature maps in the geological coordinate system, we propose spatial-wise and channel-wise feature fusion models as described in Section \ref{sec:methodology}.

\subsection{Object Detection}

The fused features $F_{fusion}$ are fed into a SSD \cite{ssd} that can regress the 3D bounding boxes and predict the confidence scores for the detected object classes. The output is a $H \times W$ feature map with $(c+7) \times B$ channels. For each of the $B$ anchor boxes, we predict $c$ class confidence scores and 7 offsets $(x, y, z, w, l, h, \theta)$ to the ground truths. 

\subsection{Loss}

The loss function in \cite{pointpillars} consists of focal loss (Eq. \ref{eq:loss1}) \cite{retinanet} for classification, and smooth $l1$ loss (Eq. \ref{eq:loss2}) for regression. The complete loss function is given below:
\begin{equation}
\begin{split}
L &= \beta_{cls}L_{cls} + \beta_{reg}L_{reg}\\
     &= \beta_{cls}L_{focal}(p) + \beta_{reg}smooth_{L_{1}}(sin(q-y_{reg}))
\end{split}
\label{eq:loss}
\end{equation}

\noindent where $\beta_{cls}=1$ and $\beta_{reg}=2$ are the classification loss and regression loss coefficients respectively, $p$ is the prediction probability, $q$ is the number of anchor boxes and $y_{reg}$ is the number of ground truth boxes.
\begin{equation}
L_{focal}(p) = -\alpha(1-p)^{\gamma}log(p)
\label{eq:loss1}
\end{equation}

\noindent where $\alpha=0.25$ and $\gamma=2$ are the parameters of the focal loss.
\begin{equation}
smooth_{L_{1}}(x) = 
    \begin{cases}
      0.5x^{2} & \text{if $|x|<1$}\\
      |x|-0.5 & \text{otherwise}
    \end{cases}       
\label{eq:loss2}
\end{equation}

\section{Proposed Adaptive Feature Fusion Models}  \label{sec:methodology}

\begin{figure*}[htbp]
\centering
\begin{subfigure}{0.95\textwidth}
\centering
    \begin{subfigure}[b]{0.3\textwidth}
    \centering
        \includegraphics[width=\textwidth]{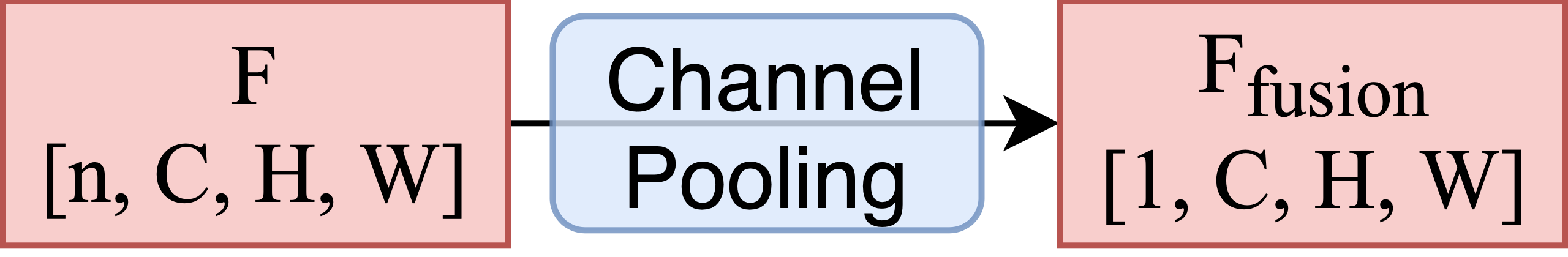}
        \caption{Spatial Fusion with channel-wise pooling such as Max pooling (MaxFusion) and Average pooling (AvgFusion) over the channel axis.}
        \label{fig:features1}
    \end{subfigure}
    \hfill
    \begin{subfigure}[b]{0.69\textwidth}
    \centering
        \includegraphics[width=0.95\textwidth]{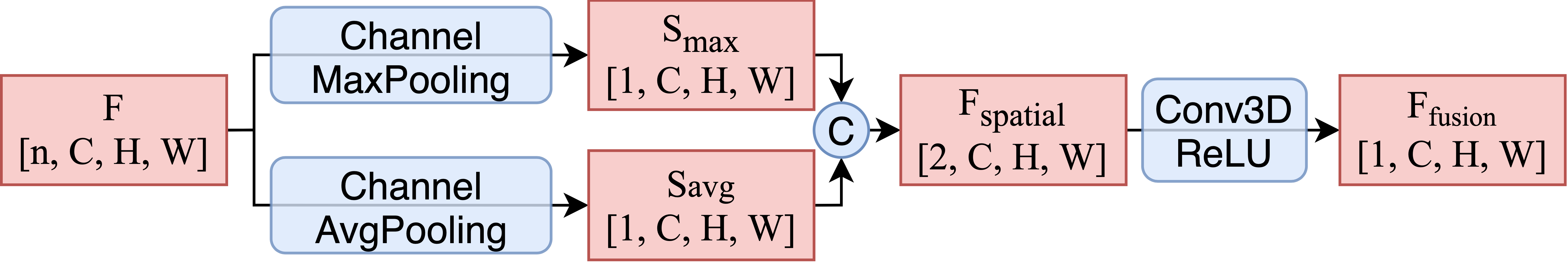}
        \caption{Spatial-wise Adaptive feature Fusion (S-AdaFusion).\newline\newline}
        \label{fig:features2}
    \end{subfigure}
\end{subfigure}
\vskip\baselineskip
\begin{subfigure}{0.95\textwidth}
\centering
    \begin{subfigure}[b]{0.3\textwidth}
    \centering
        \includegraphics[width=\textwidth]{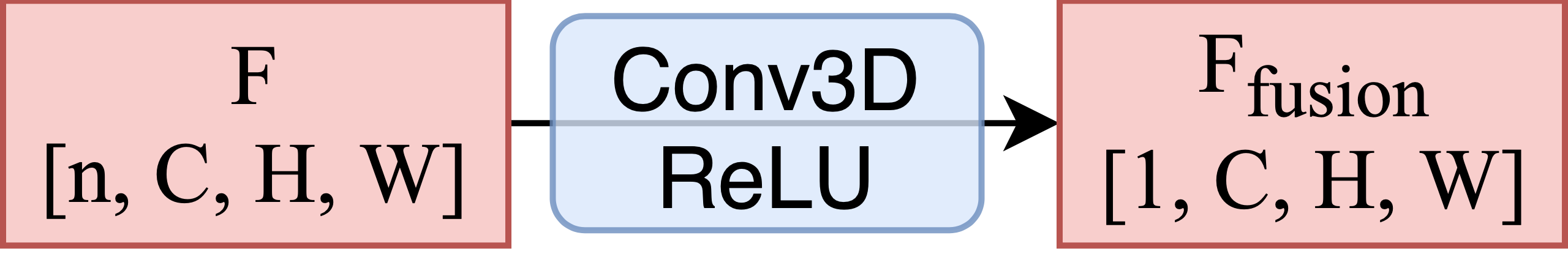}
        \caption{Channel Fusion with 3D convolution (C-3DFusion).}
        \label{fig:featurec1}
    \end{subfigure}
    \hfill
    \begin{subfigure}[b]{0.69\textwidth}
    \centering
        \includegraphics[width=0.95\textwidth]{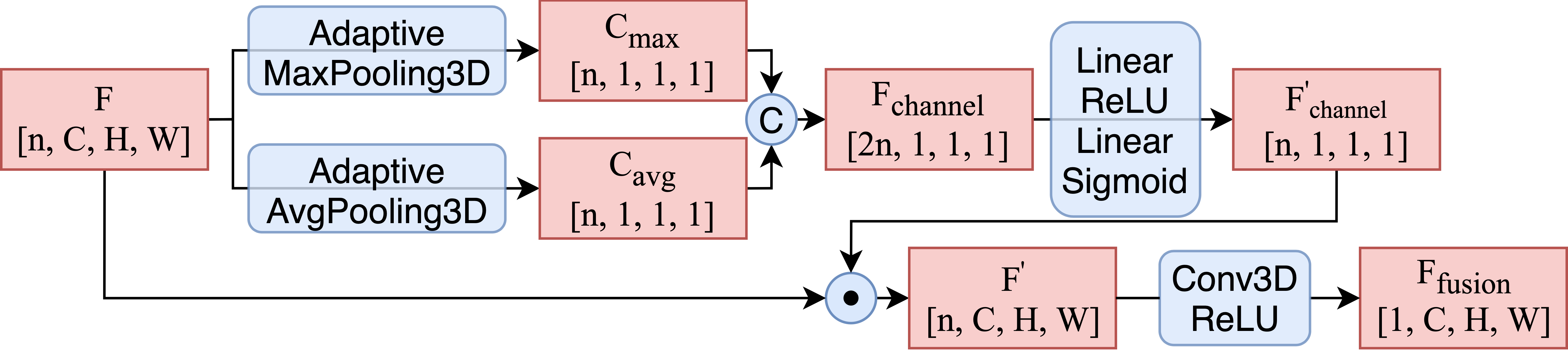}
        \caption{Channel-wise Adaptive feature Fusion (C-AdaFusion).\newline}
        \label{fig:featurec2}
    \end{subfigure}
\end{subfigure}
\caption{Feature fusion networks for intermediate feature maps aggregation.}
\label{fig:features}
\end{figure*}

In this study, we focus on intermediate feature fusion for cooperative perception and specifically on optimizing the feature fusion module to achieve better performance and higher accuracy on 3D object detection. We hypothesize that a trainable neural network can select the features more effectively than the plain reduction operators. Our proposed feature fusion models are split into spatial-wise feature fusion and channel-wise feature fusion.

\paragraph{Spatial-wise Feature Fusion.} To fuse feature maps, a straightforward reduction operator such as Max \cite{fcooper} or Mean are applied to the overlapping features as shown in Fig. \ref{fig:features1}. We refer to these two fusion methods as MaxFusion and AvgFusion in this paper, which compute max pooling and average pooling respectively, over the channel axis to get the fused feature map $F_{fusion} \in R^{1 \times C \times H \times W}$.

We propose Spatial-wise Adaptive feature Fusion (S-AdaFusion) that adaptively utilizes the spatial features generated by max pooling and average pooling as shown in Fig. \ref{fig:features2}. First, the input feature map $F \in R^{n \times C \times H \times W}$ is decomposed into $S_{max} \in R^{1 \times C \times H \times W}$ and $S_{avg} \in R^{1 \times C \times H \times W}$ by calculating max pooling and average pooling along the first channel axis. The two feature maps are concatenated together to get a 4D tensor $F_{spatial} \in R^{2 \times C \times H \times W}$ which contains two kinds of spatial information from the original concatenated intermediate feature map. Then, a 3D convolution with ReLU activation function is utilized for further feature selection and dimension reduction with the numbers of input channels and output channels equal to 2 and 1 respectively.

\paragraph{Channel-wise Feature Fusion.} CNNs perform very well in extracting features from the representations and reducing their dimensions. For an input 4D tensor $F \in R^{n \times C \times H \times W}$, a 3D CNN can be used to extract the channel features and reduce the number of input feature channels as shown in Fig. \ref{fig:featurec1} (C-3DFusion). The number of input channels of the 3D CNN would be equal to the maximum number of CAVs with a single output channel representing a combined feature set.

Inspired by the channel attention module SENet \cite{senet}, we propose a Channel-wise Adaptive feature Fusion (C-AdaFusion) module that can select and fuse the intermediate feature maps by using the channel information as shown in Fig. \ref{fig:featurec2}. Global pooling is utilized to squeeze the global information in a channel-wise descriptor. We leverage 3D adaptive max pooling and average pooling to extract two channel descriptors $C_{max} \in R^{n \times 1 \times 1 \times 1}$ and $C_{avg} \in R^{n \times 1 \times 1 \times 1}$. Then, the two vectors are concatenated and passed through two linear layers having ReLU and Sigmoid activation functions respectively. The input feature map $F \in R^{n \times C \times H \times W}$ channel-wise multiplies the learned channel weights $F^{'}_{channel} \in R^{n \times 1 \times 1 \times 1}$ to generate a new feature representation $F^{'} \in R^{n \times C \times H \times W}$. The fused feature map $F_{fusion} \in R^{1 \times C \times H \times W}$ is obtained by using a channel reduction 3D CNN.

\section{Experiments} \label{sec:experiments}

\begin{table*}[t]
\caption{Evaluation results on the OPV2V Default CARLA Towns test set for vehicle detection, OPV2V Culver City for domain adaptation, and CODD dataset for vehicle and pedestrian detection. We compare our proposed models with the baseline models and the SOTA fusion models of cooperative perception.}
\centering
\begin{tabular}{m{0.05\columnwidth}|m{0.3\columnwidth}|m{0.1\columnwidth}m{0.1\columnwidth}|m{0.1\columnwidth}m{0.1\columnwidth}||m{0.1\columnwidth}m{0.1\columnwidth}|m{0.15\columnwidth}||m{0.15\columnwidth}}
\hline

\multirow{3}{*}{\textbf{}} &
\multirow{3}{*}{\textbf{Method}} & \multicolumn{4}{c||}{\textbf{OPV2V}} & \multicolumn{3}{c||}{\textbf{CODD}} & \\

& & \multicolumn{2}{c|}{\textbf{Default Towns}} & \multicolumn{2}{c||}{\textbf{Culver City}} & \multicolumn{2}{c|}{\textbf{Vehicle}} & \multicolumn{1}{c||}{\textbf{Pedestrian}} &
\multicolumn{1}{c}{\textbf{\# Parameters}}\\

&\multirow{2}{*}{} & AP@.5 & AP@.7 & AP@.5 & AP@.7 & AP@.5 & AP@.7 & AP@.1 & \multicolumn{1}{c}{(Million)}\\
\hline
\parbox[t]{0.05\columnwidth}{\multirow{3}{*}{\rotatebox[origin=c]{90}{Baseline}}} & No Fusion & 67.9 & 60.2 & 55.7 & 47.1 & 61.7 & 55.2 & 23.6 & 6.58\\
& Early Fusion & 89.1 & 80.0 & 82.9 & 69.6 & 73.9 & 68.2 & 32.2 & 6.58\\
& Late Fusion & 85.8 & 78.1 & 79.9 & 66.8 & 66.6 & 61.7 & 27.2 & 6.58\\
\hline
\parbox[t]{0.05\columnwidth}{\multirow{4}{*}{\rotatebox[origin=c]{90}{SOTA}}} & F-Cooper \cite{fcooper} & 88.7 & 79.0 & 84.6 & 72.8 & 77.6 & 74.3 & 32.8 & 7.27\\
& AttFuse \cite{opv2v} & 90.8 & 81.5 & 85.4 & 73.5 & 81.4 & 77.7 & 38.0 & 6.58\\
& V2VNet \cite{v2vnet} & 89.7 & 82.2 & 86.0 & 73.4 & 80.3 & 75.8 & 32.0 & 14.61\\
& V2X-ViT \cite{v2xvit} & 89.1 & 82.6 & 87.3 & 73.7 & 82.3 & 78.9 & 33.8 & 13.45\\
\hline
\parbox[t]{0.05\columnwidth}{\multirow{4}{*}{\rotatebox[origin=c]{90}{Ours}}} & AvgFusion & 84.3 & 74.7 & 80.9 & 68.0 & 75.3 & 65.0 & 28.1 & 7.27\\
& C-3DFusion & 90.8 & 83.6 & 87.0 & 75.7 & 82.2 & 80.1 & 39.5 & 7.27\\
& C-AdaFusion & 88.5 & 81.4 & 85.9 & 72.4 & 83.4 & 80.8 & 37.5 & 7.27\\
& S-AdaFusion & \textbf{91.6} & \textbf{85.6} & \textbf{88.1} & \textbf{79.0} & \textbf{86.2} & \textbf{83.9} & \textbf{45.2} & 7.27\\
\hline
\end{tabular}
\label{table:results}
\end{table*}

We conduct experiments on the publicly available cooperative perception datasets OPV2V dataset \cite{opv2v} and the CODD dataset \cite{codd}. We compare the prediction accuracy for different perception tasks to validate the performance of the proposed models against the conventional single vehicle perception model (no fusion) and multiple cooperative perception benchmark models.

\subsection{Datasets}

\paragraph{OPV2V Dataset.} The OPV2V dataset \cite{opv2v} is built with OpenCDA simulation tool \cite{opencda}. It has two subsets, a Default CARLA Towns and a Culver City. The Default CARLA Towns contains 6,765 training samples, 1,980 validation samples, and 2,170 testing samples in eight CARLA default towns. The Culver City contains 550 samples to test the domain adaptation ability of the model. The number of CAVs in this dataset ranges between [2, 7], and each CAV has its LiDAR information and labeled 3D bounding boxes.

\paragraph{CODD Dataset.} The CODD dataset \cite{codd} is built with CARLA simulation tool \cite{carla} and contains 108 snippets in eight CARLA towns where each snippet has 125 frames. The first 100 frames are used for training and the last 25 frames are used for testing. The numbers of vehicles and pedestrians of this dataset range between [4, 15] and [2, 8]. The dataset contains all vehicles' LiDAR information and labeled vehicle and pedestrian 3D bounding boxes, and this is the only cooperative perception dataset that contains the pedestrian class so far.

\subsection{Implementation Details}

During training, a random group of CAVs are selected from the scene with a defined upper limit of CAVs including an ego vehicle. For validation purposes, the ego vehicle and the CAVs are fixed for a fair comparison.

To compare with other benchmarks, we follow the parameter settings in \cite{opv2v}. The ranges of $x, y, z$ are [(-140.8, 140.8), (-40, 40), (-3, 1)] meters on the OPV2V dataset. We set the $x, y, z$ ranges to [(-140.8, 140.8), (-40, 40), (-6, 4)] meters on the CODD dataset. The $x$ and $y$ resolutions of the pillar are $0.4$ meters in both datasets. The vehicle anchor and pedestrian anchor have a (length, width, height) of (3.9, 1.6, 1.56) meters and (0.6, 0.6, 1.7) meters respectively.

Our model is implemented using the PyTorch framework, trained and evaluated on Tesla V100 GPU having 32 GB RAM, CUDA v10.1 and cuDNN v9.1. Early stopping, multi-step scheduler, and Adam optimizer with an $\epsilon$ of 0.1 and a weight decay of 0.0001 are used to train the network.

At inference stage, we use a confidence score threshold of 0.2. Non-maximum Suppression (NMS) with an Intersection-over-Union (IoU) threshold of 0.15 is applied to remove the redundant predictions. The effectiveness of the model is evaluated with the common metric Average Precision (AP) of vehicle with 0.5 and 0.7 IoU thresholds (AP@0.5 and AP@0.7) as well as the AP of pedestrian with 0.1 IoU threshold (AP@0.1).

\subsection{Results}

\begin{figure*}[t]
\centering
\begin{subfigure}{0.9\textwidth}
\centering
    \begin{subfigure}{0.45\textwidth}
        \includegraphics[width=\textwidth]{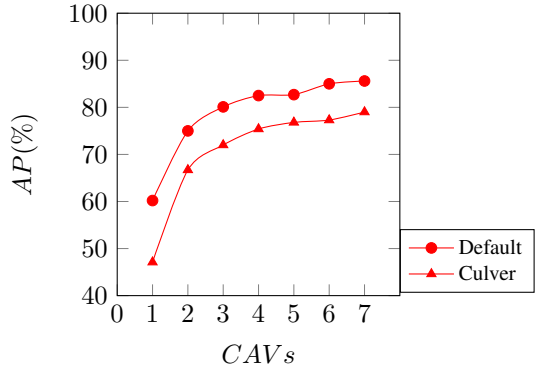}
        \caption{Cooperative perception results on OPV2V dataset.}
        \label{fig:coopcood}
    \end{subfigure}
    \hfill
    \begin{subfigure}{0.45\textwidth}
        \includegraphics[width=\textwidth]{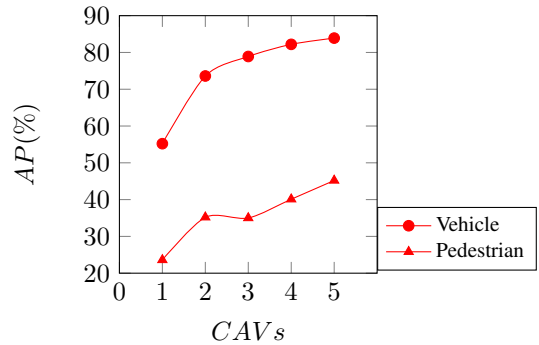}
        \caption{Cooperative perception results on CODD dataset.}
        \label{fig:coopcodd}
    \end{subfigure}
 \end{subfigure}
\caption{The influence of the number of CAVs for Cooperative perception with S-AdaFusion model architecture.}
\label{fig:coopresults}
\end{figure*}

The evaluation results on the OPV2V and CODD datasets are shown in Table \ref{table:results}. We compare our proposed models with the baseline models including no fusion, early fusion and late fusion, and multiple SOTA fusion models of cooperative perception \cite{fcooper,v2vnet,v2xvit,opv2v}. Fig. \ref{fig:coopresults} displays the cooperative perception results with different number of CAVs. When the number of CAVs is one, it is the traditional perception process. The maximum number of CAVs is seven and five on the OPV2V and CODD datasets respectively.

\begin{figure*}[htbp]
\centering
\begin{subfigure}{0.95\textwidth}
\centering
    \begin{subfigure}{0.49\textwidth}
        \includegraphics[width=\textwidth]{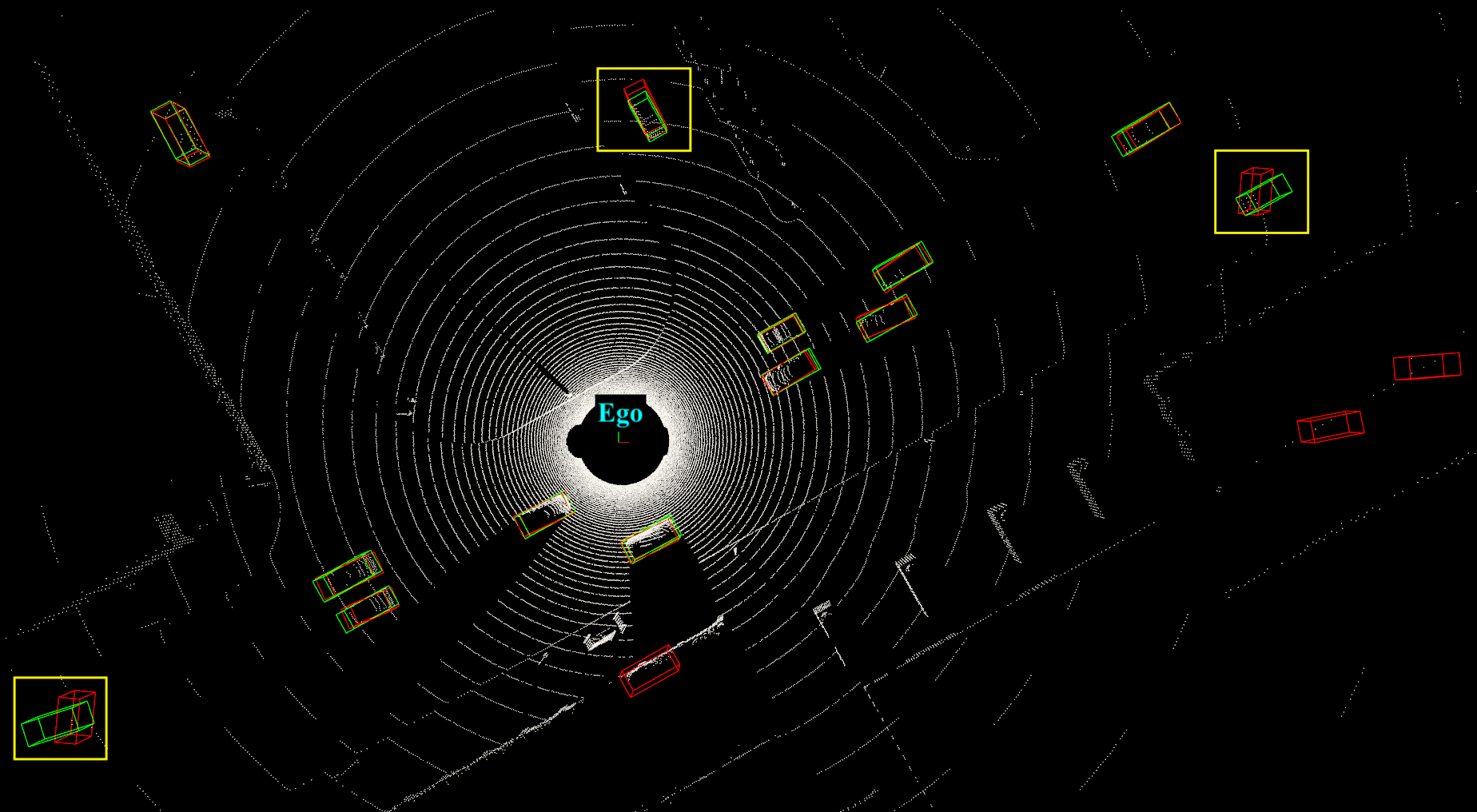}
    \end{subfigure}
    \hfill
    \begin{subfigure}{0.49\textwidth}
        \includegraphics[width=\textwidth]{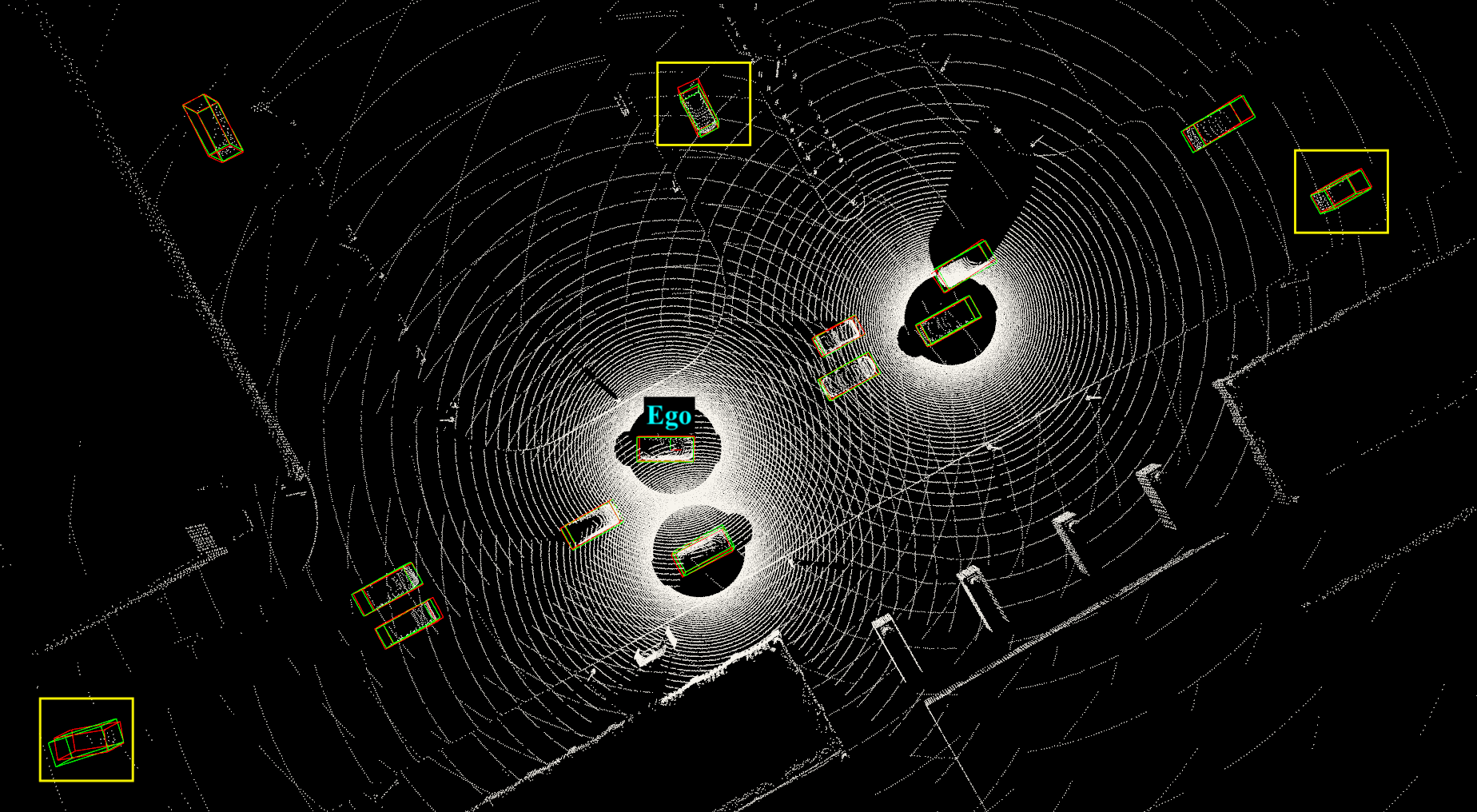}
    \end{subfigure}
    \begin{subfigure}{0.49\textwidth}
        \includegraphics[width=\textwidth]{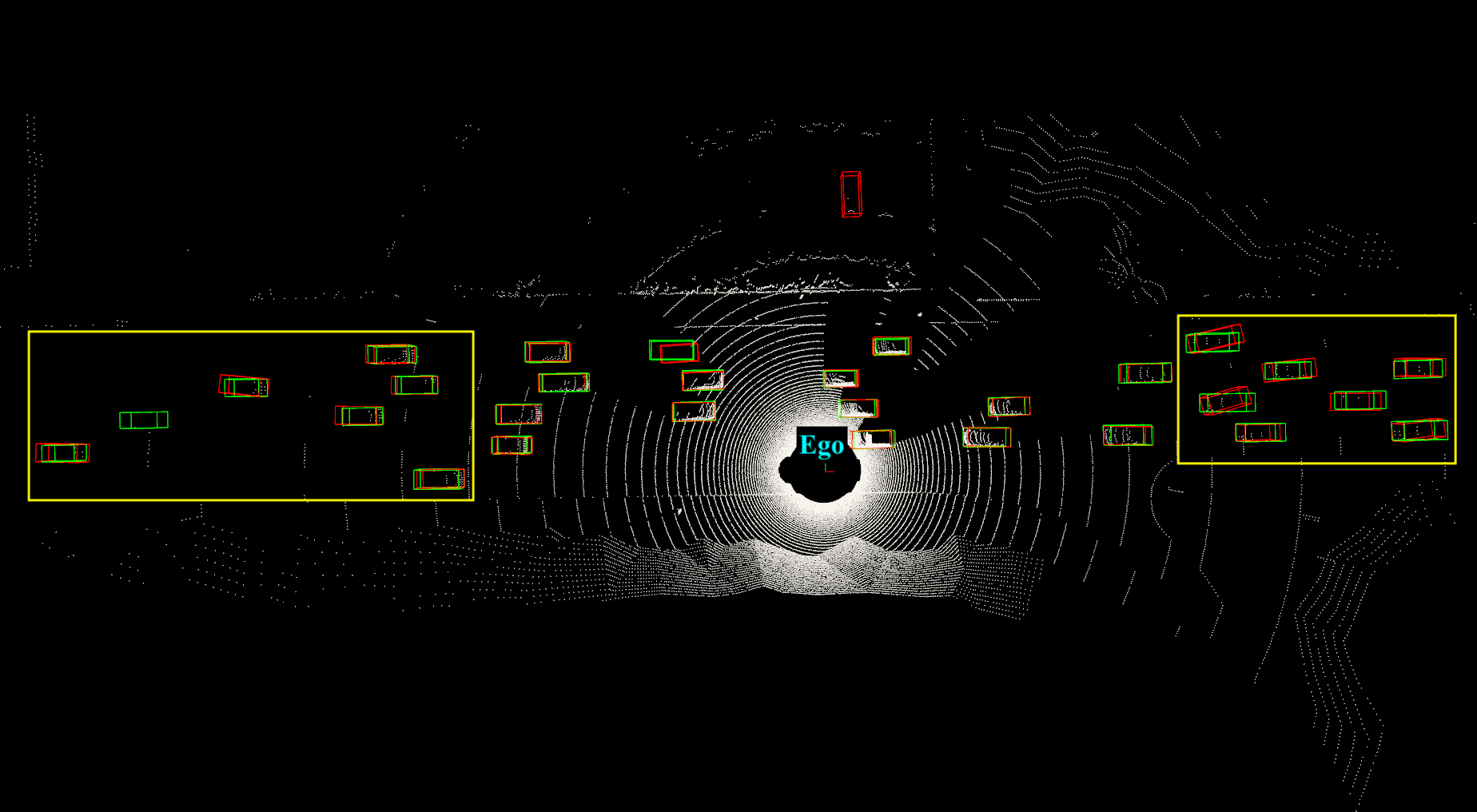}
        \caption{Single vehicle perception (no fusion).}
        \label{fig:vsa}
    \end{subfigure}
    \hfill
    \begin{subfigure}{0.49\textwidth}
        \includegraphics[width=\textwidth]{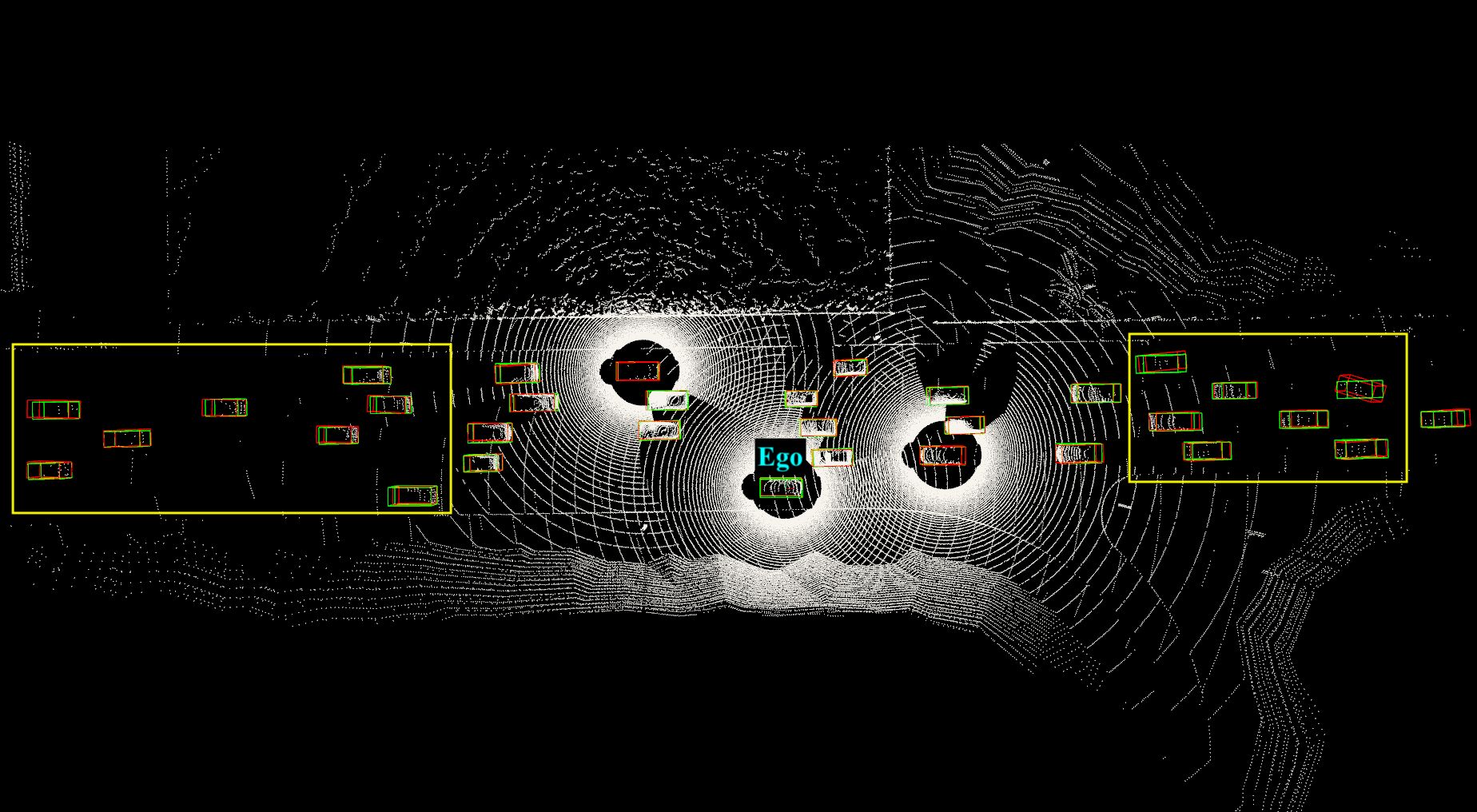}
        \caption{Cooperative perception with intermediate fusion.}
        \label{fig:vsb}
    \end{subfigure}
\end{subfigure}
\caption{Examples of single vehicle perception (no fusion) and cooperative perception on OPV2V dataset. The ground truth and predicted 3D bounding boxes are depicted in green and red respectively. Yellow rectangles highlight the vehicles that are occluded or further away from the ego vehicle.}
\label{fig:vs}
\end{figure*}

\begin{figure}[htbp]
\centering
\includegraphics[width=0.99\columnwidth]{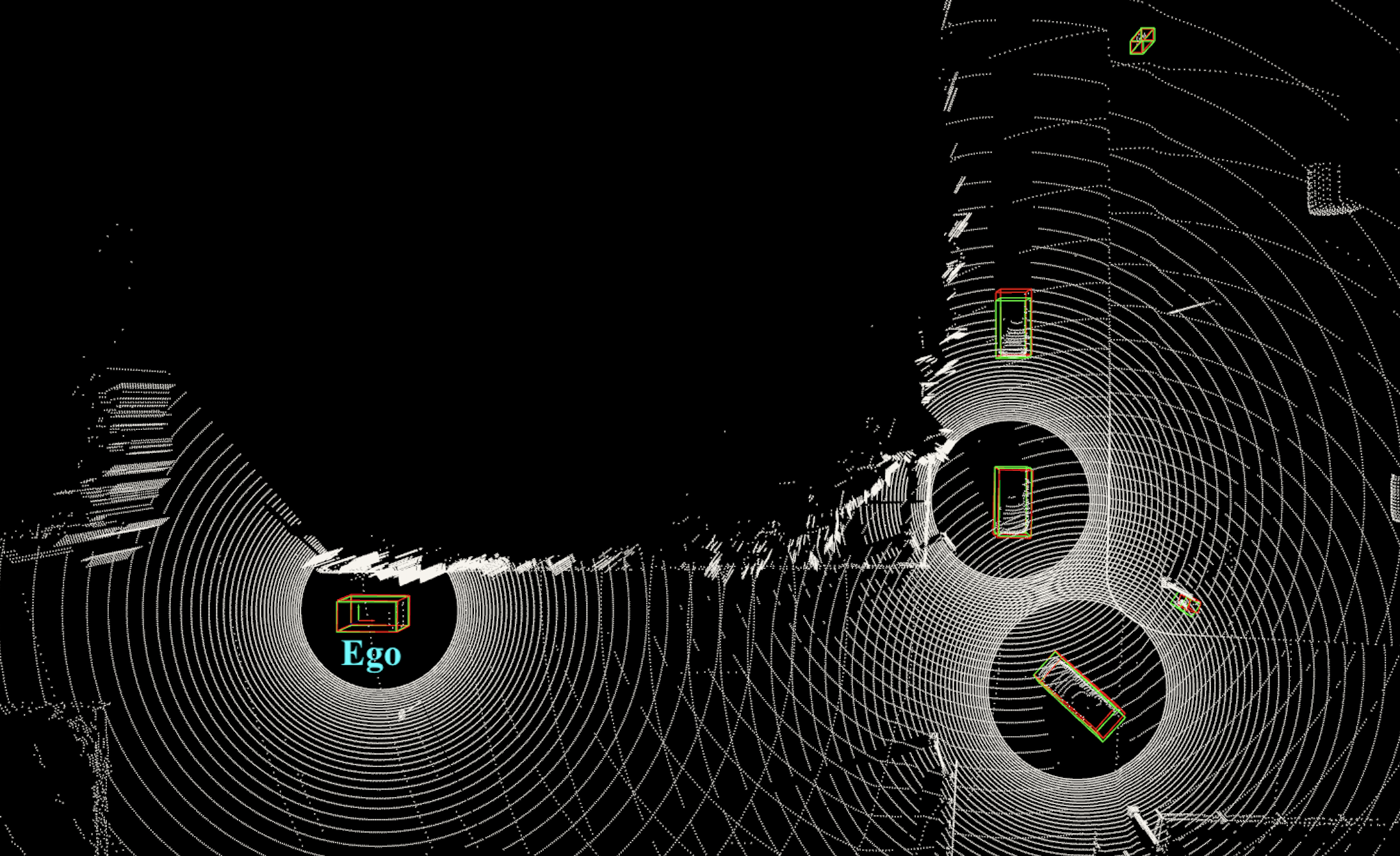}
\caption{An example of solving occlusion situation with cooperative perception.}
\label{fig:problem}
\end{figure}

Some prediction outputs from our S-AdaFusion are shown in Fig. \ref{fig:vs} and Fig. \ref{fig:problem}. Without cooperative perception, the no fusion model incorrectly identifies bushes and constructions as vehicles and generates more false positives as shown in Fig. \ref{fig:vsa}. Comparing the predictions in the yellow rectangles, cooperative perception is more robust than no fusion in regressing the bounding boxes for the vehicles further away from the ego vehicle. Regardless of whether the vehicles are occluded, other CAVs can assist the ego vehicle to perceive occluded objects and get a better overall perception result. Fig. \ref{fig:problem} shows that with the assistance of the other two CAVs, the ego vehicle can still detect the vehicle and pedestrian even when they are fully occluded on the vertical street.

\subsection{Discussion}

From Table \ref{table:results} we can find that most of the intermediate fusion models outperform the early and late fusion. Our S-AdaFusion outperforms all the models on all the perception tasks including vehicle detection, pedestrian detection and domain adaptation. The SOTA models and our proposed adaptive feature fusion models all achieve around or over 90\% AP@0.5 on OPV2V Default CARLA Towns set, whereas our S-AdaFusion model obtains 85.6\% AP@0.7 which is the highest among the existing models. This means even if all the models can achieve high precision in classification, our model is much better on the 3D bounding box regression. The S-AdaFusion model achieves more significant improvements in domain adaptation which is at least 5.3\% higher in AP@0.7 than other models. The large margin of improvement in domain adaptation illustrates the advantage of using pooling and the generalization of our proposed models in feature fusion.

The evaluation results on the CODD dataset further validate the effectiveness of our proposed models on both vehicle and pedestrian detection. Our C-AdaFusion outperforms other models for vehicle detection. Our S-AdaFusion model surpasses other benchmark models by at least 3.9\% AP@0.5 and 5.0\% AP@0.7 for vehicle detection, and 7.2\% AP@0.1 for pedestrian detection which are impressive improvements. Additionally, our architecture has fewer parameters than current SOTA V2VNet and V2X-ViT.

Fig. \ref{fig:coopresults} demonstrates that vehicle detection accuracy and domain adaptation accuracy are both improved along with the increasing number of CAVs. The accuracy of pedestrian detection is positively correlated to the number of CAVs. The cooperative perception improves over 20\% AP@0.7 on vehicle detection and over 20\% AP@0.1 on pedestrian detection comparing to the traditional perception (no fusion). Although cooperative perception can enhance pedestrian detection accuracy, it does not achieve the accuracy we expected. We presume two possible reasons led to this observation. First, the number of pedestrians in CODD dataset is low and the distribution is sparse which make the detection very difficult. Second, the model has limitations in small object detection. As a result, the margin of improvement is not as significant as vehicle detection.




\begin{table}[t]
\caption{Ablation studies of proposed models with different kernel sizes on the OPV2V dataset.}
\centering
\begin{tabular}{m{0.35\columnwidth}|m{0.1\columnwidth}m{0.1\columnwidth}|m{0.1\columnwidth}m{0.1\columnwidth}}
\hline
\multirow{2}{*}{\textbf{Method}} & \multicolumn{2}{c|}{\textbf{Default}} & \multicolumn{2}{c}{\textbf{Culver}}\\
\multirow{2}{*}{} & AP@.5 & AP@.7 & AP@.5 & AP@.7\\
\hline
C-3DFusion (ks=1) & 83.7 & 73.6 & 79.2 & 63.0\\
C-3DFusion (ks=3) & 90.8 & 83.6 & 87.0 & 75.7\\
S-AdaFusion (ks=1) & 91.8 & 83.0 & 86.1 & 70.4\\
S-AdaFusion (ks=3) & 91.6 & 85.6 & 88.1 & 79.0\\
\hline
\end{tabular}
\footnotesize{\raggedleft "ks" stands for kernel size \par}
\label{table:ablation}
\end{table}

\subsection{Ablation Study}

The F-Cooper \cite{fcooper} (MaxFusion), AvgFusion, and our proposed C-3DFusion and S-AdaFusion are used in a set of experiments conducted for an ablation study. As shown in Table \ref{table:results}, the F-Cooper method applying MaxFusion model achieves much better results than the AvgFusion model, since it selects the most distinctive features from multiple feature maps. Without using the spatial information, the C-3DFusion model with trainable 3D convolution layer can select and fuse features from multiple CAVs. Additionally, switching the kernel from $1\times1\times1$ channel reduction kernel (stride=1, padding=0) to $3\times3\times3$ kernel (stride=1, padding=1) enhances the accuracy by about 10\% AP@0.7 on vehicle detection as shown in Table \ref{table:ablation}. This is because the larger kernel not only focus on one cell and reduce the channel, it can also increase the perception field and extract some spatial features from the neighboring cells. It is interesting to note that the C-3DFusion already surpasses all other models in most experiments. By incorporating these modules, S-AdaFusion, the adaptive feature selection and fusion model, achieves the highest AP on all the validation datasets for cooperative perception.

\section{Conclusion} \label{sec:conclusion}

With the advancements in sensing technology for AVs and VC systems, novel techniques are being explored to address the challenges in vehicular perception especially in traffic object detection. In this research, we study cooperative perception using LiDAR point cloud data to address some limitations of the conventional object detection process such as sensor resolution and object occlusion. We propose adaptive feature fusion models with trainable neural network for intermediate fusion of cooperative perception and validate the superiority of the models through a number of experiments using two benchmark datasets. The research validates the hypothesis that a trainable neural network module for intermediate feature fusion can improve the object detection accuracy in cooperative perception. Our S-AdaFusion model outperforms the SOTA intermediate feature fusion cooperative perception models on the two benchmark datasets.

To improve the accuracy of pedestrian detection, a different point encoding or object detection network can be explored to improve the pedestrian detection accuracy. Additionally, images can include more human features which may be helpful for pedestrian and cyclist detection.

{\small
\bibliographystyle{ieee_fullname}
\bibliography{egpaper}

\begin{thebibliography}{10}\itemsep=-1pt

\bibitem{infrastructurecooper}
Eduardo Arnold, Mehrdad Dianati, Robert de Temple, and Saber Fallah.
\newblock Cooperative perception for 3d object detection in driving scenarios
  using infrastructure sensors.
\newblock {\em IEEE Transactions on Intelligent Transportation Systems}, 2020.

\bibitem{codd}
Eduardo Arnold, Sajjad Mozaffari, and Mehrdad Dianati.
\newblock {Fast and Robust Registration of Partially Overlapping Point Clouds}.
\newblock 2021.

\bibitem{fcooper}
Qi Chen, Xu Ma, Sihai Tang, Jingda Guo, Qing Yang, and Song Fu.
\newblock F-cooper: Feature based cooperative perception for autonomous vehicle
  edge computing system using 3d point clouds.
\newblock In {\em Proceedings of the 4th ACM/IEEE Symposium on Edge Computing},
  pages 88--100, 2019.

\bibitem{cooper}
Qi Chen, Sihai Tang, Qing Yang, and Song Fu.
\newblock Cooper: Cooperative perception for connected autonomous vehicles
  based on 3d point clouds.
\newblock In {\em 2019 IEEE 39th International Conference on Distributed
  Computing Systems (ICDCS)}, pages 514--524. IEEE, 2019.

\bibitem{voxelrcnn}
Jiajun Deng, Shaoshuai Shi, Peiwei Li, Wengang Zhou, Yanyong Zhang, and
  Houqiang Li.
\newblock Voxel r-cnn: Towards high performance voxel-based 3d object
  detection.
\newblock In {\em Proceedings of the AAAI Conference on Artificial
  Intelligence}, volume~35, pages 1201--1209, 2021.

\bibitem{carla}
Alexey Dosovitskiy, German Ros, Felipe Codevilla, Antonio Lopez, and Vladlen
  Koltun.
\newblock Carla: An open urban driving simulator.
\newblock In {\em Conference on robot learning}, pages 1--16. PMLR, 2017.

\bibitem{kitti}
Andreas Geiger, Philip Lenz, and Raquel Urtasun.
\newblock Are we ready for autonomous driving? the kitti vision benchmark
  suite.
\newblock In {\em 2012 IEEE conference on computer vision and pattern
  recognition}, pages 3354--3361. IEEE, 2012.

\bibitem{senet}
Jie Hu, Li Shen, and Gang Sun.
\newblock Squeeze-and-excitation networks.
\newblock In {\em Proceedings of the IEEE conference on computer vision and
  pattern recognition}, pages 7132--7141, 2018.

\bibitem{trupercept}
Braden Hurl, Robin Cohen, Krzysztof Czarnecki, and Steven Waslander.
\newblock Trupercept: Trust modelling for autonomous vehicle cooperative
  perception from synthetic data.
\newblock In {\em 2020 IEEE Intelligent Vehicles Symposium (IV)}, pages
  341--347. IEEE, 2020.

\bibitem{pointpillars}
Alex~H Lang, Sourabh Vora, Holger Caesar, Lubing Zhou, Jiong Yang, and Oscar
  Beijbom.
\newblock Pointpillars: Fast encoders for object detection from point clouds.
\newblock In {\em Proceedings of the IEEE/CVF Conference on Computer Vision and
  Pattern Recognition}, pages 12697--12705, 2019.

\bibitem{bevfusion2}
Tingting Liang, Hongwei Xie, Kaicheng Yu, Zhongyu Xia, Zhiwei Lin, Yongtao
  Wang, Tao Tang, Bing Wang, and Zhi Tang.
\newblock Bevfusion: A simple and robust lidar-camera fusion framework.
\newblock {\em arXiv preprint arXiv:2205.13790}, 2022.

\bibitem{fpn}
Tsung-Yi Lin, Piotr Doll{\'a}r, Ross Girshick, Kaiming He, Bharath Hariharan,
  and Serge Belongie.
\newblock Feature pyramid networks for object detection.
\newblock In {\em Proceedings of the IEEE conference on computer vision and
  pattern recognition}, pages 2117--2125, 2017.

\bibitem{retinanet}
Tsung-Yi Lin, Priya Goyal, Ross Girshick, Kaiming He, and Piotr Doll{\'a}r.
\newblock Focal loss for dense object detection.
\newblock In {\em Proceedings of the IEEE international conference on computer
  vision}, pages 2980--2988, 2017.

\bibitem{ssd}
Wei Liu, Dragomir Anguelov, Dumitru Erhan, Christian Szegedy, Scott Reed,
  Cheng-Yang Fu, and Alexander~C Berg.
\newblock Ssd: Single shot multibox detector.
\newblock In {\em European conference on computer vision}, pages 21--37.
  Springer, 2016.

\bibitem{bevfusion1}
Zhijian Liu, Haotian Tang, Alexander Amini, Xinyu Yang, Huizi Mao, Daniela Rus,
  and Song Han.
\newblock Bevfusion: Multi-task multi-sensor fusion with unified bird's-eye
  view representation.
\newblock {\em arXiv preprint arXiv:2205.13542}, 2022.

\bibitem{sumcooper}
Ehsan~Emad Marvasti, Arash Raftari, Amir~Emad Marvasti, Yaser~P Fallah, Rui
  Guo, and Hongsheng Lu.
\newblock Cooperative lidar object detection via feature sharing in deep
  networks.
\newblock In {\em 2020 IEEE 92nd Vehicular Technology Conference
  (VTC2020-Fall)}, pages 1--7. IEEE, 2020.

\bibitem{pointnet}
Charles~R Qi, Hao Su, Kaichun Mo, and Leonidas~J Guibas.
\newblock Pointnet: Deep learning on point sets for 3d classification and
  segmentation.
\newblock In {\em Proceedings of the IEEE conference on computer vision and
  pattern recognition}, pages 652--660, 2017.

\bibitem{qiao2020dilated}
Donghao Qiao and Farhana Zulkernine.
\newblock Dilated squeeze-and-excitation u-net for fetal ultrasound image
  segmentation.
\newblock In {\em 2020 IEEE Conference on Computational Intelligence in
  Bioinformatics and Computational Biology (CIBCB)}, pages 1--7. IEEE, 2020.

\bibitem{qiao2021drivable}
Donghao Qiao and Farhana Zulkernine.
\newblock Drivable area detection using deep learning models for autonomous
  driving.
\newblock In {\em 2021 IEEE International Conference on Big Data (Big Data)},
  pages 5233--5238. IEEE, 2021.

\bibitem{pvrcnn++}
Shaoshuai Shi, Li Jiang, Jiajun Deng, Zhe Wang, Chaoxu Guo, Jianping Shi,
  Xiaogang Wang, and Hongsheng Li.
\newblock Pv-rcnn++: Point-voxel feature set abstraction with local vector
  representation for 3d object detection.
\newblock {\em arXiv preprint arXiv:2102.00463}, 2021.

\bibitem{attention}
Ashish Vaswani, Noam Shazeer, Niki Parmar, Jakob Uszkoreit, Llion Jones,
  Aidan~N Gomez, {\L}ukasz Kaiser, and Illia Polosukhin.
\newblock Attention is all you need.
\newblock {\em Advances in neural information processing systems}, 30, 2017.

\bibitem{v2vnet}
Tsun-Hsuan Wang, Sivabalan Manivasagam, Ming Liang, Bin Yang, Wenyuan Zeng, and
  Raquel Urtasun.
\newblock V2vnet: Vehicle-to-vehicle communication for joint perception and
  prediction.
\newblock In {\em European Conference on Computer Vision}, pages 605--621.
  Springer, 2020.

\bibitem{cbam}
Sanghyun Woo, Jongchan Park, Joon-Young Lee, and In~So Kweon.
\newblock Cbam: Convolutional block attention module.
\newblock In {\em Proceedings of the European conference on computer vision
  (ECCV)}, pages 3--19, 2018.

\bibitem{opencda}
Runsheng Xu, Yi Guo, Xu Han, Xin Xia, Hao Xiang, and Jiaqi Ma.
\newblock Opencda: an open cooperative driving automation framework integrated
  with co-simulation.
\newblock In {\em 2021 IEEE International Intelligent Transportation Systems
  Conference (ITSC)}, pages 1155--1162. IEEE, 2021.

\bibitem{cobevt}
Runsheng Xu, Zhengzhong Tu, Hao Xiang, Wei Shao, Bolei Zhou, and Jiaqi Ma.
\newblock Cobevt: Cooperative bird's eye view semantic segmentation with sparse
  transformers.
\newblock {\em arXiv preprint arXiv:2207.02202}, 2022.

\bibitem{v2xvit}
Runsheng Xu, Hao Xiang, Zhengzhong Tu, Xin Xia, Ming-Hsuan Yang, and Jiaqi Ma.
\newblock V2x-vit: Vehicle-to-everything cooperative perception with vision
  transformer.
\newblock {\em arXiv preprint arXiv:2203.10638}, 2022.

\bibitem{opv2v}
Runsheng Xu, Hao Xiang, Xin Xia, Xu Han, Jinlong Liu, and Jiaqi Ma.
\newblock Opv2v: An open benchmark dataset and fusion pipeline for perception
  with vehicle-to-vehicle communication.
\newblock {\em arXiv preprint arXiv:2109.07644}, 2021.

\bibitem{second}
Yan Yan, Yuxing Mao, and Bo Li.
\newblock Second: Sparsely embedded convolutional detection.
\newblock {\em Sensors}, 18(10):3337, 2018.

\bibitem{3dcvf}
Jin~Hyeok Yoo, Yecheol Kim, Jisong Kim, and Jun~Won Choi.
\newblock 3d-cvf: Generating joint camera and lidar features using cross-view
  spatial feature fusion for 3d object detection.
\newblock In {\em European Conference on Computer Vision}, pages 720--736.
  Springer, 2020.

\bibitem{zhang2021distributed}
Zijian Zhang, Shuai Wang, Yuncong Hong, Liangkai Zhou, and Qi Hao.
\newblock Distributed dynamic map fusion via federated learning for intelligent
  networked vehicles.
\newblock In {\em 2021 IEEE International Conference on Robotics and Automation
  (ICRA)}, pages 953--959. IEEE, 2021.

\bibitem{voxelnet}
Yin Zhou and Oncel Tuzel.
\newblock Voxelnet: End-to-end learning for point cloud based 3d object
  detection.
\newblock In {\em Proceedings of the IEEE conference on computer vision and
  pattern recognition}, pages 4490--4499, 2018.

\end{thebibliography}
}

\end{document}